\documentclass[sigconf]{acmart}
\AtBeginDocument{%
  }

\copyrightyear{2025}
\acmYear{2025}
\setcopyright{acmlicensed}
\acmConference[MM '25] {Proceedings of the 33rd ACM International Conference on Multimedia}{October 27--31, 2025}{Dublin, Ireland.}
\acmBooktitle{Proceedings of the 33rd ACM International Conference on Multimedia (MM '25), October 27--31, 2025, Dublin, Ireland}
\acmISBN{979-8-4007-2035-2/2025/10}
\acmDOI{10.1145/3746027.3755285}

\usepackage{xcolor}
\usepackage{colortbl} 
\usepackage{graphicx}
\usepackage{multirow}
\usepackage{balance}
\usepackage{makecell}

\begin{document}

\title{DisFaceRep: Representation Disentanglement for Co-occurring Facial Components in Weakly Supervised Face Parsing}

\author{Xiaoqin Wang}
\affiliation{
  \institution{School of Computer Science and Software Engineering, Shenzhen University}
  \city{Shenzhen}
  \country{China}}
\email{wangxiaoqin2022@email.szu.edu.cn}
\orcid{0000-0002-6813-7666}

\author{Xianxu Hou}
\authornotemark[1]
\affiliation{
  \institution{School of AI and Advanced Computing, Xi’an Jiaotong-Liverpool University}
  \city{Suzhou}
  \country{China}}
\email{hxianxu@gmail.com}
\orcid{0000-0002-8728-2842}

\author{Meidan Ding}
\affiliation{
  \institution{School of Computer Science and Software Engineering, Shenzhen University}
  \city{Shenzhen}
  \country{China}}
\email{dingmeidan2023@email.szu.edu.cn}
\orcid{0009-0001-0520-020X}

\author{Junliang Chen}
\affiliation{
  \institution{Department of Electrical and Electronic Engineering, The Hong Kong Polytechnic University}
  \city{Hong Kong}
  \country{China}}
\email{jun-liang.chen@connect.polyu.hk}
\orcid{0000-0001-7516-9546}

\author{Kaijun Deng}
\affiliation{%
  \institution{School of Computer Science and Software Engineering, Shenzhen University}
  \city{Shenzhen}
  \country{China}}
\email{dengkaijun2023@email.szu.edu.cn}
\orcid{0009-0009-1352-2952}

\author{Jinheng Xie}
\affiliation{
  \institution{Department of Electrical and Computer Engineering, National University of Singapore}
  \city{Singapore}
  \country{Singapore}}
\email{sierkinhane@gmail.com}
\orcid{0000-0001-5678-4500}

\author{Linlin Shen}
\authornote{Corresponding authors}
\affiliation{
  \institution{Computer Vision Institute, School of Artificial Intelligence, Shenzhen University}
  \institution{Guangdong Provincial Key Laboratory of Intelligent Information Processing, Shenzhen University}
  \city{Shenzhen}
  \country{China}}
\email{llshen@szu.edu.cn}
\orcid{0000-0003-1420-0815}

\renewcommand{\shortauthors}{Xiaoqin Wang et al.}

\begin{abstract}
Face parsing aims to segment facial images into key components such as eyes, lips, and eyebrows. While existing methods rely on dense pixel-level annotations, such annotations are expensive and labor-intensive to obtain. To reduce annotation cost, we introduce Weakly Supervised Face Parsing (WSFP), a new task setting that performs dense facial component segmentation using only weak supervision, such as image-level labels and natural language descriptions. WSFP introduces unique challenges due to the high co-occurrence and visual similarity of facial components, which lead to ambiguous activations and degraded parsing performance. To address this, we propose DisFaceRep, a representation disentanglement framework designed to separate co-occurring facial components through both explicit and implicit mechanisms. Specifically, we introduce a co-occurring component disentanglement strategy to explicitly reduce dataset-level bias, and a text-guided component disentanglement loss to guide component separation using language supervision implicitly. Extensive experiments on CelebAMask-HQ, LaPa, and Helen demonstrate the difficulty of WSFP and the effectiveness of DisFaceRep, which significantly outperforms existing weakly supervised semantic segmentation methods. The code will be released at \href{https://github.com/CVI-SZU/DisFaceRep}{\textcolor{cyan}{https://github.com/CVI-SZU/DisFaceRep}}.
\end{abstract}




\maketitle

\section{Introduction}
Face parsing, a fine-grained semantic segmentation task, aims to assign pixel-level labels to facial images by distinguishing key components such as the eyes, nose, lips, hair, and ears. Accurate facial component segmentation is critical for a wide range of downstream applications, including face swapping \cite{nirkin2019fsgan}, face editing \cite{lee2020maskgan}, face completion \cite{li2017generative}, and virtual makeup synthesis \cite{ou2016beauty}.
Thanks to the availability of densely annotated datasets, existing methods based on Convolutional Neural Networks (CNNs) \cite{zhou2015interlinked, zhou2017face, te2020edge, luo2020ehanet, liu2015multi, lin2021roi, guo2018residual} and, more recently, Transformers \cite{narayan2024segface}, have achieved impressive performance. However, creating pixel-level annotations, especially for small and intricate facial features, is time-consuming and not scalable to large datasets or diverse domains.

To alleviate the cost of dense annotations, weakly supervised learning has emerged as a compelling alternative in the semantic segmentation community, formalized as Weakly Supervised Semantic Segmentation (WSSS). It aims to predict pixel-level segmentation masks using only weak supervision signals, including image-level labels \cite{ahn2018learning, pathak2015constrained, pinheiro2015image}, bounding boxes \cite{dai2015boxsup, papandreou2015weakly}, point annotations \cite{bearman2016s}, scribbles \cite{lin2016scribblesup, vernaza2017learning}, or natural language descriptions \cite{xie2022clims}. Despite relying on coarse supervision, recent approaches \cite{jo2022recurseed, wu2024dupl, jo2023mars} have achieved performance competitive with, and in some cases exceeding, that of fully supervised methods—highlighting the remarkable potential of weak supervision for dense prediction tasks.

\begin{figure}[t]
    \centering
    {\includegraphics[width=0.90\linewidth]{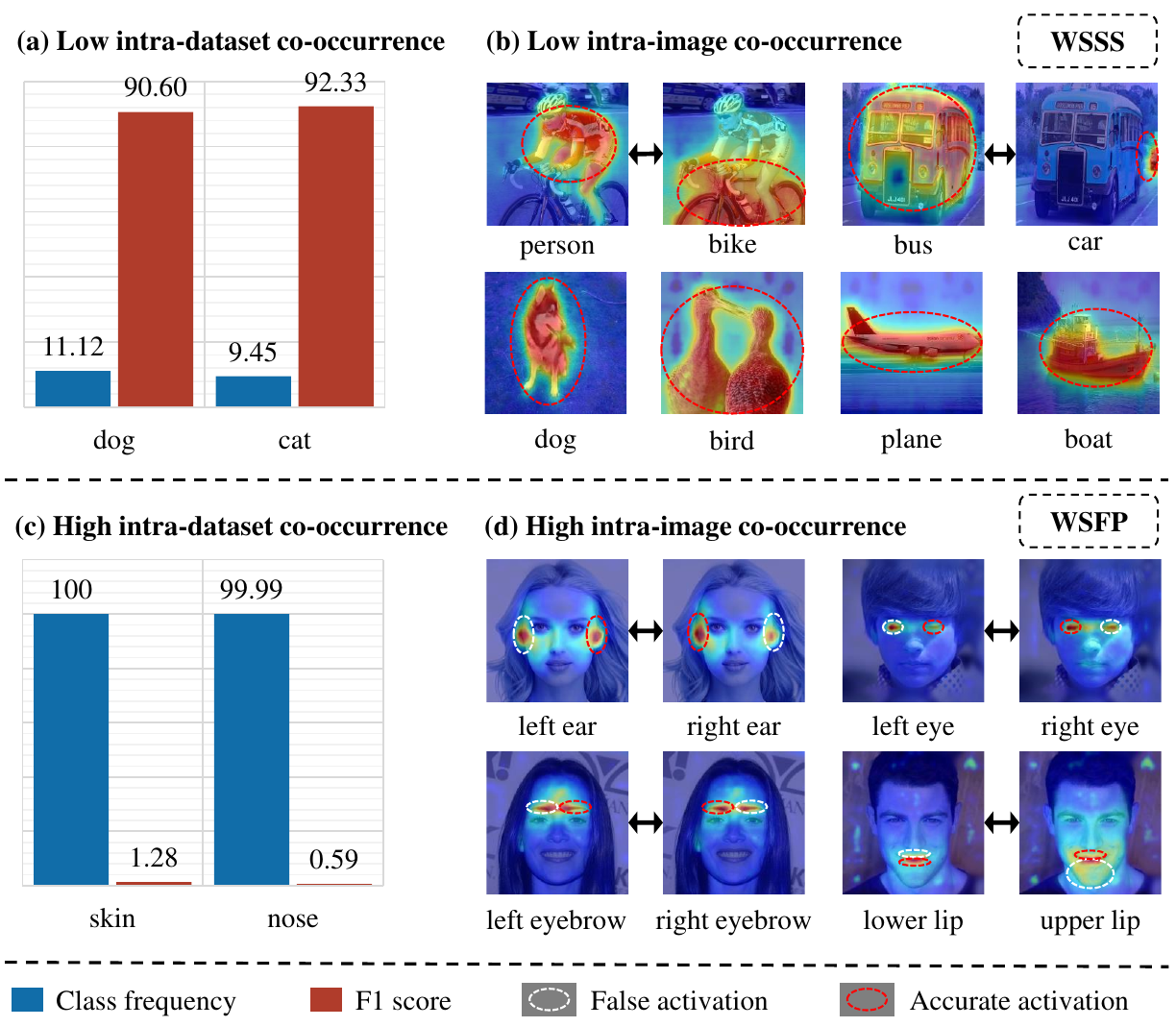}}
    \vspace{-4pt}
    \caption{Comparison of co-occurrence challenges in WSSS and WSFP. (a-b) PASCAL VOC \cite{everingham2010pascal} shows low intra-dataset and intra-image co-occurrence, along with high inter-class dissimilarity, resulting in high F1 scores. In contrast, (c–d) CelebAMask-HQ \cite{lee2020maskgan} exhibits high intra-dataset and intra-image co-occurrence of visually similar facial components, such as skin, nose, ears, eyes, eyebrows, and lips. This frequent co-occurrence and visual similarity of components lead to false activations and low F1 scores when WSSS methods \cite{yang2024separate} are directly applied.}
    \label{fig_wsfp}
    \vspace{-12pt}
\end{figure}

While weakly supervised learning has achieved notable success in general semantic segmentation, it remains largely underexplored in the context of face parsing. To bridge this gap, we propose Weakly Supervised Face Parsing (WSFP), a new task that extends the WSSS paradigm to the facial domain by enabling dense component segmentation under weak annotations. A natural baseline for WSFP is to directly apply existing WSSS methods; however, this proves suboptimal due to challenges unique to WSFP, particularly the difficulty in generating accurate Facial Component Activation Maps (FCAMs). Unlike general semantic segmentation, WSFP suffers from significant activation overlap due to the frequent co-occurrence and high visual similarity of facial components, which degrades the quality of pseudo labels. This issue manifests in two forms: (1) intra-dataset co-occurrence: datasets like CelebAMask-HQ exhibit high co-occurrence of facial categories (e.g., skin and nose appear in nearly every facial image), reducing the discriminative capacity of classification models, as shown in Fig.~\ref{fig_wsfp}(c); and (2) intra-image co-occurrence: facial components often share similar visual features (e.g., texture and shape) but differ in spatial location (e.g., left vs. right eyes), causing over-activation of visually similar but semantically unrelated regions, as shown in Fig.~\ref{fig_wsfp}(d). These two forms of co-occurrence present core challenges in WSFP and significantly limit the effectiveness of applying WSSS methods to facial data.

To address these challenges, we propose DisFaceRep (Fig.~\ref{fig:overview}), a novel representation disentanglement framework tailored to the WSFP task. DisFaceRep is built upon two complementary modules: Co-occurring Component Disentanglement (CCD) and Text-guided Component Disentanglement (TCD). CCD explicitly mitigates dataset-level bias by randomly masking dominant facial regions that frequently co-occur using Grounding DINO \cite{liu2025grounding}. This prevents the model from over-relying on high-frequency components during learning. On the other hand, TCD implicitly enhances component separation by aligning visual and textual representations through the facial language pretraining model FLIP \cite{li2024flip}. It leverages both positive and negative text-component associations to guide fine-grained activation and suppress semantically unrelated regions. By combining these two strategies, DisFaceRep generates high-quality Facial Component Activation Maps (FCAMs), significantly improving parsing performance under weak supervision. The main contributions of this work are as follows:
\begin{itemize} 
    \item We introduce WSFP, a new task that extends WSSS to facial component parsing. We identify and analyze the core challenges of frequent co-occurrence and visual similarity, which limit the effectiveness of directly applying existing WSSS's methods to WSFP.
    \item We propose DisFaceRep, a representation disentanglement framework that addresses these challenges through two complementary modules: Co-occurring Component Disentanglement, which explicitly reduces dataset-level bias, and Text-guided Component Disentanglement, which implicitly guides component separation via vision-language alignment.
    \item We conduct extensive experiments on CelebAMask-HQ, LaPa, and Helen, demonstrating that DisFaceRep significantly outperforms state-of-the-art WSSS methods in both segmentation accuracy and component-level precision.
\end{itemize}

\section{Related Works}
\subsection{Fully Supervised Face Parsing}
With the advancement of deep learning, numerous fully supervised methods have been developed for face parsing, which can be broadly categorized into global-based and local-based approaches.

Global-based methods aim to predict pixel-wise labels for the entire face image using features extracted from convolutional or graph-based networks. Early CNN-based models \cite{jackson2016cnn, wei2017learning, zhou2017face, lin2019face, liu2015multi} directly learn from RGB inputs to capture spatial relationships among facial regions. More recent efforts incorporate structural priors: AGRNet \cite{te2021agrnet} and EAGRNet \cite{te2020edge} utilize graph representations and edge information to enhance spatial consistency, while DML-CSR \cite{zheng2022decoupled} combines dual graph convolutional networks with cyclic self-regulation to handle noisy labels. FP-LIIF \cite{sarkar2023parameter} proposes a local implicit function to exploit facial symmetry and continuity. In addition, FaRL \cite{zheng2022general} explores large-scale vision-language pretraining on facial data to boost downstream tasks such as face parsing.

Local-based methods, by contrast, decompose the parsing task into separate sub-networks, each dedicated to a specific facial region. Luo et al. \cite{luo2012hierarchical} adopt a hierarchical network that segments each facial part individually. Zhou et al. \cite{zhou2015interlinked} propose an interlinked CNN model that performs post-detection pixel classification, albeit with high memory and computation cost. Liu et al. \cite{liu2017face} introduce a two-stage method combining a shallow CNN and spatial RNN to balance accuracy and inference speed. Despite the remarkable progress of both paradigms, they all rely heavily on dense pixel-level labels, which are expensive and time-consuming to obtain.

\subsection{Weakly Supervised Semantic Segmentation}
Weakly supervised semantic segmentation aims to reduce annotation costs by training models with weak supervision, such as image-level labels. The standard WSSS pipeline typically follows a three-stage process: (1) a classification network is trained using weak labels to generate Class Activation Maps (CAMs) \cite{zhou2016learning}; (2) these initial CAMs are refined into pseudo masks using post-processing methods such as dense CRF \cite{krahenbuhl2011efficient}, pixel affinity propagation \cite{ahn2018learning, ahn2019weakly}, or saliency-based techniques \cite{jiang2019integral, liu2020leveraging}; and (3) the refined pseudo masks are used to supervise a segmentation network.

Due to the discriminative nature of classification models, CAMs typically highlight only the most salient object regions. To address this limitation, various methods have been proposed to enhance CAM completeness. These include cross-image attention \cite{li2021group, sun2020mining}, adversarial erasing \cite{kweon2021unlocking, kweon2023weakly, yoon2022adversarial, zhang2018adversarial}, consistency regularization through data augmentation \cite{wang2020self, zhang2021complementary}, boundary-aware refinement \cite{fan2020learning, kolesnikov2016seed, rong2023boundary}, loss function modification \cite{chen2022class, wu2022adaptive}, and Transformer-based approaches \cite{rossetti2022max, xu2022multi, xu2024mctformer+}. Recently, vision-language pretraining, particularly CLIP, has gained traction in WSSS. CLIMS \cite{xie2022clims} first introduced CLIP to improve object region activation. CLIP-ES \cite{lin2023clip} enhances CAM quality using text prompts and GradCAM \cite{selvaraju2017grad}, while WeakCLIP \cite{zhu2024weakclip} proposes a fine-grained text-to-pixel alignment paradigm. WeCLIP \cite{zhang2024frozen} further explores a single-stage approach leveraging a frozen CLIP backbone for CAM refinement.

These methods demonstrate the effectiveness of weak supervision for semantic segmentation in general domains, particularly by improving CAM quality. However, face parsing under weak supervision remains largely underexplored. In this work, we analyze the unique challenges of weakly supervised learning in the face parsing setting and introduce a novel task and framework tailored for low-cost, fine-grained facial segmentation.

\begin{figure*}[t]
    \centering
    {\includegraphics[width=0.92\linewidth]{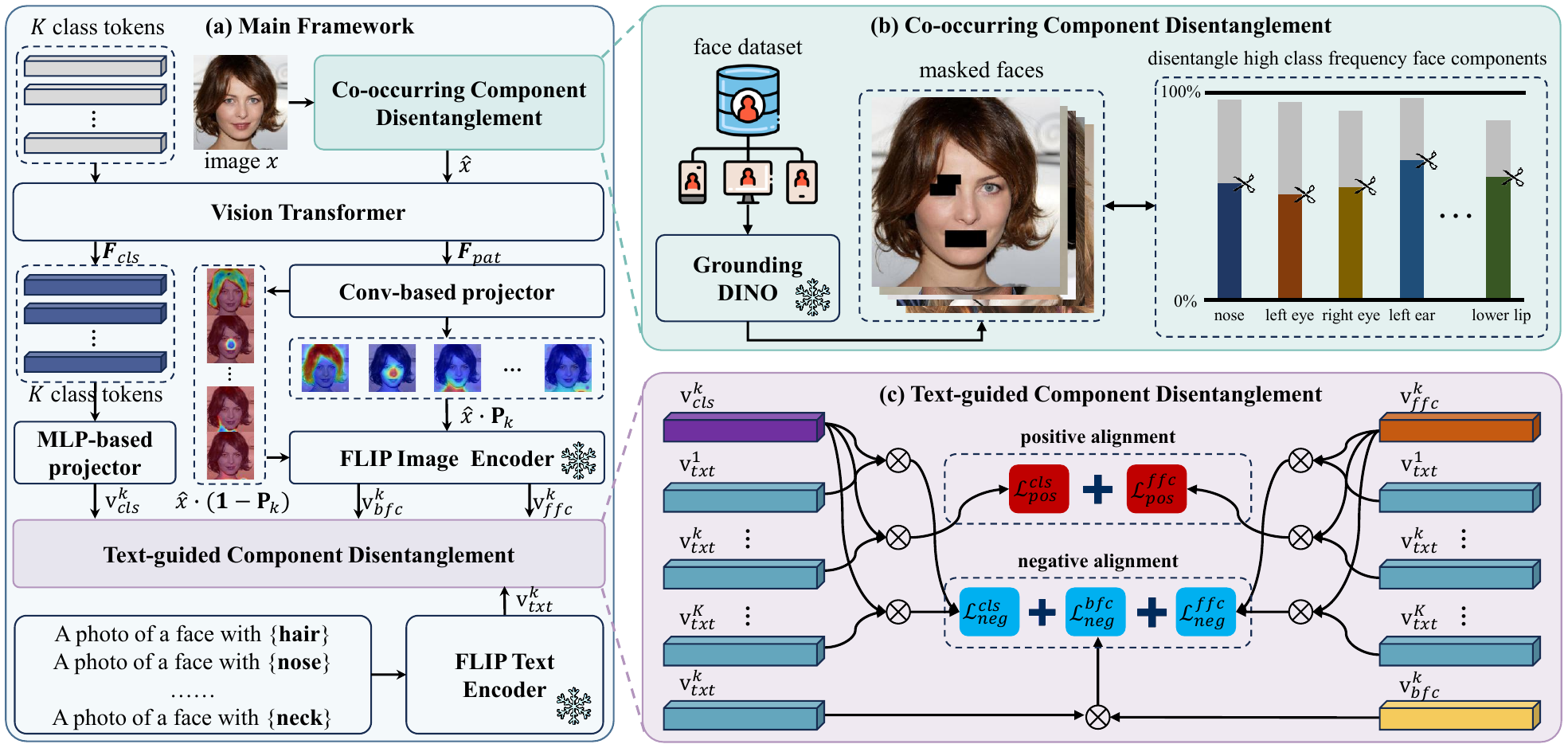}}
    \vspace{-4pt}
    \caption{An overview of our proposed DisFaceRep for WSFP. (a) The main framework for predicting initial FCAMs. $\mathbf{v}_{cls}^k$, $\mathbf{v}_{ffc}^k$, $\mathbf{v}_{bfc}^k$, and $\mathbf{v}_{txt}^k$ are the $k$-th representations corresponding to the class token, foreground facial component, background facial component, and text description, respectively. $\hat{x}\cdot\mathbf{P}_k$ and $\hat{x}\cdot\left(1-\mathbf{P}_k\right)$ are FCAMs for the foreground and background, respectively. (b) Co-occurring component disentanglement explicitly suppresses dominant facial components via the frozen Grounding DINO \cite{liu2024grounding} model. (c) Text-guided component disentanglement implicitly separates component representations via positive alignment ( $\mathcal{L}_{pos}^{cls}$ and $\mathcal{L}_{pos}^{ffc}$) and negative alignment ( $\mathcal{L}_{neg}^{cls}$, $\mathcal{L}_{neg}^{ffc}$, and $\mathcal{L}_{neg}^{bfc}$). $\bigotimes$ denotes cosine similarity.}
    \label{fig:overview}
    \vspace{-6pt}
\end{figure*}

\section{Methodology}
\subsection{Problem Definition and Key Challenges}
In this section, we formally define the Weakly Supervised Face Parsing (WSFP) setting. Given a training set $\mathcal{D}=\{x_i, y_i\}_{i=1}^{n}$, where $x_i \in \mathcal{X}$ represents the $i$-th face image and $y_i \in \mathcal{Y}$ denotes its corresponding image-level label, the objective of WSFP is to segment a face image $x$ into $K$ facial components using only the weak supervisory signal $y$. In contrast to the WSSS paradigm, the WSFP pipeline consists of two stages: first, a face component classification model is trained to generate Facial Component Activation Maps (FCAMs) under weak supervision, and then pseudo masks are generated by setting a threshold $\theta$ on FCAMs; second, these pseudo masks are used as dense supervision to train a face parsing model.

Among the two stages in the WSFP pipeline, generating accurate and complete FCAMs is the most critical and challenging step. A key obstacle is the strong co-occurrence among facial components, which hinders the model’s ability to distinguish and localize them accurately. This challenge arises in two main forms. First, there is intra-dataset co-occurrence, where certain facial components co-occur in a large majority of training images. As shown in Fig.\ref{fig_wsfp}(c), components such as skin and nose appear in over 99\% of images in the dataset, leading to high class frequency and poor discriminability across components. Second, the human face exhibits strong intra-image co-occurrence, with components that have similar appearance but distinct spatial positions (e.g., left/right eyes and eyebrows). As illustrated in Fig.\ref{fig_wsfp}(d), traditional classification networks tend to over-activate such components, resulting in ambiguous localization. These two forms of co-occurrence and visual similarity significantly degrade the quality of FCAMs, thereby limiting the performance of face parsing. To address this, we propose DisFaceRep, a representation disentanglement framework designed to resolve component entanglement under weak supervision.

\subsection{Overview of DisFaceRep}
The overall framework of DisFaceRep is depicted in Fig.~\ref{fig:overview}. Given a training set $\mathcal{D}=\{x_i, y_i\}_{i=1}^{n}$, our method first applies Co-occurring Component Disentanglement (CCD) to suppress high-frequency facial components explicitly. This is achieved by randomly masking regions associated with frequently co-occurring attributes, resulting in a new training set $\hat{\mathcal{D}} = \{\hat{x}_i, \hat{y}_i\}_{i=1}^{n}$, where $\hat{x}_i$ and $\hat{y}_i$ denote the images and labels with reduced co-occurrence bias. Following \cite{xu2024mctformer+}, we adopt a multi-class token ViT~\cite{dosovitskiy2020image} as the backbone, and feed $\hat{x}$ into it to extract class tokens $\mathbf{F}_{cls} \in \mathbb{R}^{K \times d}$ and patch tokens $\mathbf{F}_{pat} \in \mathbb{R}^{N \times d}$, where $K$ and $N$ represent the number of categories and image patches, and $d$ is the embedding dimension. An MLP-based projector transforms the class tokens into a vision-language space, while patch tokens are reshaped and passed through a convolutional layer and a sigmoid function. The resulting feature map $\mathbf{P}_k$ is element-wise multiplied with $\hat{x}$ to extract foreground facial component feature $\mathbf{v}_{ffc}^k$, while background component feature $\mathbf{v}_{bfc}^k$ is computed by multiplying $\left(1-\mathbf{P}_k\right)$ with $\hat{x}$. To further enhance component separation, we introduce Text-guided Component Disentanglement (TCD), which leverages FLIP \cite{li2024flip}—a face-specific vision-language model built upon CLIP~\cite{radford2021learning}—for implicitly aligning visual features with semantic concepts. Finally, we extract accurate and complete FCAMs from the trained classification network following the procedure in \cite{xu2024mctformer+}.

\subsection{Co-occurring Component Disentanglement}
Formally, CCD leverages the Grounding DINO \cite{liu2024grounding}, which is an open-set object detection method, to randomly mask some facial components based on predicted bounding boxes. Due to the presence of multiple similar facial components, relying solely on the highest-confidence bounding box for coordinate prediction may yield inaccurate results. To solve this problem, we design a multi-threshold filtering strategy to obtain a set of accurate prediction boxes. Specifically, there is a set of candidate coordinates $B^i = \left[h_1^i, w_1^i, h_2^i, w_2^i\right],  i \in \{1,2, \dots, m\}$ with the highest confidence for a set of facial components that are masked out. The condition for the accurate prediction box of the $i$-th masked component is:
\begin{equation}
    \label{eq1}
    C^i = 
    \begin{cases}
         \text{True} \ \cap \ \left(\dfrac{w_1^i + w_2^i}{2} < \dfrac{W}{2}\right), & \text{if right components} \\
         \text{True} \ \cap \ \left(\dfrac{w_1^i + w_2^i}{2} > \dfrac{W}{2}\right), & \text{if left components} \\
         \text{True}, & \text{otherwise} \\
    \end{cases}
\end{equation}
where $W$ is the width of the image. The pixel coordinate position set of the face components with high category frequency is $M=\bigcup_{i=1}^{m}\{B_i | C_i\}$. Finally, we can accurately mask out some components with high class frequency as follows:
\begin{equation}
    \label{eq1}
    \hat{x}(i,j) = 
    \begin{cases}
         0, & \text{if} \ (i,j) \in M \\
         x(i,j), & \text{otherwise}
    \end{cases}
\end{equation}
where $x(i,j)$ is the pixel value of the face image $x$. The $\hat{y}$ corresponds to the image label of $\hat{x}$. In this work, the candidate set of facial components that may be masked out in the image is ``nose, left/right eye, left/right eyebrow, left/right ear, and mouth". These components are high-frequency, cover common face parsing datasets, and have no overlap between boxes of different components. Therefore, masking out in the facial images can mitigate dataset-level bias.

\subsection{Text-guided Component Disentanglement}
\label{TCD}
After applying CCD, we obtain a new training set $\hat{\mathcal{D}} = \{\hat{x}_i, \hat{y}_i\}_{i=1}^{n}$ with reduced high class frequency and lower co-occurrence bias. As shown in Fig.~\ref{fig:overview}, the $k$-th class token and $\hat{x}$ are input to the ViT backbone, and then obtain the class embedding $\mathbf{F}_{cls}^k$ and patch embedding $\mathbf{F}_{pat}^k$. Then, $\mathbf{F}_{cls}^k$ is projected into the class representation $\mathbf{v}_{cls}^k$ via an MLP-based projector:
\begin{equation}
    \mathbf{v}_{cls}^k=\mathbf{W}^T\mathbf{F}_{cls}^k
    \label{eq1}
\end{equation}
where $\mathbf{W}$ is projection matrix. Thus, each class representation vector corresponds to distinct facial semantic information.

In addition, a convolutional layer and a sigmoid function $\sigma(\cdot)$ are directly applied to the patch token $\mathbf{F}_{pat}^k$ and generate the patch attention map $\mathbf{P}_k$:
\begin{equation}
    \mathbf{P}_k=\sigma\left(\text{conv}\left(\mathbf{F}_{pat}^k\right)\right)
    \label{eq1}
\end{equation}

To enable text-guided facial component disentanglement, we use the image encoder $\varepsilon_v(\cdot)$ from the facial domain CLIP, i.e., FLIP ~\cite{li2024flip} to generate both the foreground and background facial component representation vectors:
\begin{equation}
    \mathbf{v}_{ffc}^k=\varepsilon_v\left(\hat{x}\cdot\mathbf{P}_k\right)
    \label{eq1}
\end{equation}
\begin{equation}
    \mathbf{v}_{bfc}^k=\varepsilon_v\left(\hat{x}\cdot\left(1-\mathbf{P}_k\right)\right)
    \label{eq1}
\end{equation}
where $\mathbf{v}_{ffc}^k$ is the $k$-th foreground facial component representation vector, and $\mathbf{v}_{bfc}^k$ is the $k$-th background facial component representation vector.

To disentangle entangled facial components using text supervision, we construct a text prompt $T_k$, i.e., `a photo of a face with \{component\}', e.g., `a photo of a face with nose', following FLIP \cite{li2024flip}. Then, the text encoder $\varepsilon_t(\cdot)$ from FLIP is utilized to obtain the text representation vector $\mathbf{v}_{txt}^k$ of the $k$-th text prompt:
\begin{equation}
    \mathbf{v}_{txt}^k=\varepsilon_t\left(T_k\right)
    \label{eq1}
\end{equation}

Finally, based on the $\mathbf{v}_{cls}^k$, $\mathbf{v}_{ffc}^k$, $\mathbf{v}_{bfc}^k$, and $\mathbf{v}_{txt}^k$, the proposed TCD implicitly disentangles different facial components through the following positive and negative alignment losses.

\subsubsection{Positive Alignment}
It aims to pull facial component features close to the corresponding text representations. Specifically, given the $k$-th class representation $\mathbf{v}_{cls}^k$ and its corresponding text representation $\mathbf{v}_{txt}^k$, we begin by calculating the cosine similarity between $\mathbf{v}_{cls}^k$ and $\mathbf{v}_{txt}^k$, and then the positive loss in terms of class representation $\mathcal{L}_{pos}^{cls}$ is:
\begin{equation}
    \label{eq-pos-cls} 
    \mathcal{L}_{pos}^{cls} = -\sum_{k=1}^{K}\hat{y}_k \cdot \log\left(\mathbf{sim}\left(\mathbf{v}_{cls}^k, \mathbf{v}_{txt}^k\right)\right)
\end{equation}
where $\mathbf{sim} \left(\cdot\right)$ indicates the cosine similarity.

To improve the completeness of activated component regions, we also encourage the model to maintain the feature consistency of $\mathbf{v}_{ffc}^k$ and $\mathbf{v}_{txt}^k$. Thus, we design the foreground component and the text prompt matching loss $\mathcal{L}_{pos}^{ffc}$ as:
\begin{equation}
    \label{eq-pos-fg}
    \mathcal{L}_{pos}^{ffc} = -\sum_{k=1}^{K}\hat{y}_k \cdot \log\left(\mathbf{sim}\left(\mathbf{v}_{ffc}^k, \mathbf{v}_{txt}^k\right)\right)
\end{equation}

Thus, the positive alignment consists of $\mathcal{L}_{pos}^{cls}$ and $\mathcal{L}_{pos}^{ffc}$ as:
\begin{equation}
\label{eq: pos}
\mathcal{L}_{pos} = \alpha_0\mathcal{L}_{pos}^{cls} + \alpha_1\mathcal{L}_{pos}^{ffc}
\end{equation}
where $\alpha_0$ and $\alpha_1$ are weighting coefficients. By maximizing $\mathcal{L}_{pos}$, the model is encouraged to focus on discriminative regions associated with each facial component.

\subsubsection{Negative Alignment}
Due to frequently co-occurring and visually similar components, the classification model trained only with $\mathcal{L}_{pos}$ often over-activates non-target regions. To mitigate this, we introduce three negative losses that explicitly suppress spurious activations and promote better component separation.

Specifically, given the class representation $\mathbf{v}_{cls}^k$ and the $l$-th non-corresponding text representation $\mathbf{v}_{txt}^{k,l}$, the $\mathcal{L}_{neg}^{cls}$ is used to suppress the co-occurring components in the class representation space as:
\begin{equation}
    \label{cls-neg}
    \mathcal{L}_{neg}^{cls} = -\sum_{k=1}^{K}\sum_{l=1}^{L}\hat{y}_k \cdot \log\left(1 - \mathbf{sim}\left(\mathbf{v}_{cls}^k, \mathbf{v}_{txt}^{k,l}\right)\right)
\end{equation}
To further suppress co-occurring facial components, we design the $\mathcal{L}_{neg}^{ffc}$ loss for disentangling the $\mathbf{v}_{ffc}^k$ from the $l$-th non-corresponding text representation $\mathbf{v}_{txt}^{k,l}$:
\begin{equation}
\label{eq1}
\mathcal{L}_{neg}^{ffc} = -\sum_{k=1}^{K}\sum_{l=1}^{L}\hat{y}_k \cdot \log\left(1 - \mathbf{sim}\left(\mathbf{v}_{ffc}^k, \mathbf{v}_{txt}^{k,l}\right)\right)
\end{equation}
We also introduce the background region and text representation matching loss $\mathcal{L}_{neg}^{bfc}$ to  account for broader contextual information and reduce false activations:
\begin{equation}
    \label{eq1}
    \mathcal{L}_{neg}^{bfc} = -\sum_{k=1}^{K}\hat{y}_k \cdot \log\left(1 - \mathbf{sim}\left(\mathbf{v}_{bfc}^k, \mathbf{v}_{txt}^k\right)\right)
\end{equation}
Finally, we combine the above negative losses as follows: 
\begin{equation}
    \label{eq: neg}
    \mathcal{L}_{neg} = \beta_0\mathcal{L}_{neg}^{cls} + \beta_1\mathcal{L}_{neg}^{ffc} + \beta_2\mathcal{L}_{neg}^{bfc}
\end{equation}
where $\beta_0$, $\beta_1$, and $\beta_2$ are weighting coefficients. Minimizing $\mathcal{L}_{neg}$ encourages the model to suppress irrelevant activations and better disentangle complete and distinct facial components.

\subsubsection{Activation Regularization}
Following \cite{xie2022clims}, DisFaceRep incorporates the activation regularization loss for ensuring that the irrelevant backgrounds are excluded in the activation map $\mathbf{P}_k$:
\begin{equation}
    \mathcal{L}_{reg} = \frac{1}{K}\sum_{k=1}^{K}S_k, \ \ S_k = \frac{1}{HW}\sum_{h=1}^{H}\sum_{w=1}^{W}\mathbf{P}_k(h,w)
    \label{eq_reg_cam}
\end{equation}
where $S_k$ is the activation area of the $k$-th facial component, and $H$ is the height of the image.

\subsubsection{Overall Objective}
The overall training objective of the proposed DisFaceRep framework integrates both alignment and regularization losses and is defined as:
\begin{equation}
\label{eq: all}
\mathcal{L} = \mathcal{L}_{pos} + \mathcal{L}_{neg} + \lambda\mathcal{L}_{reg}
\end{equation}
where $\lambda$ is the hyper-parameter weighting the three loss terms.

\begin{figure*}[t]
    \centering
    {\includegraphics[width=0.90\linewidth]{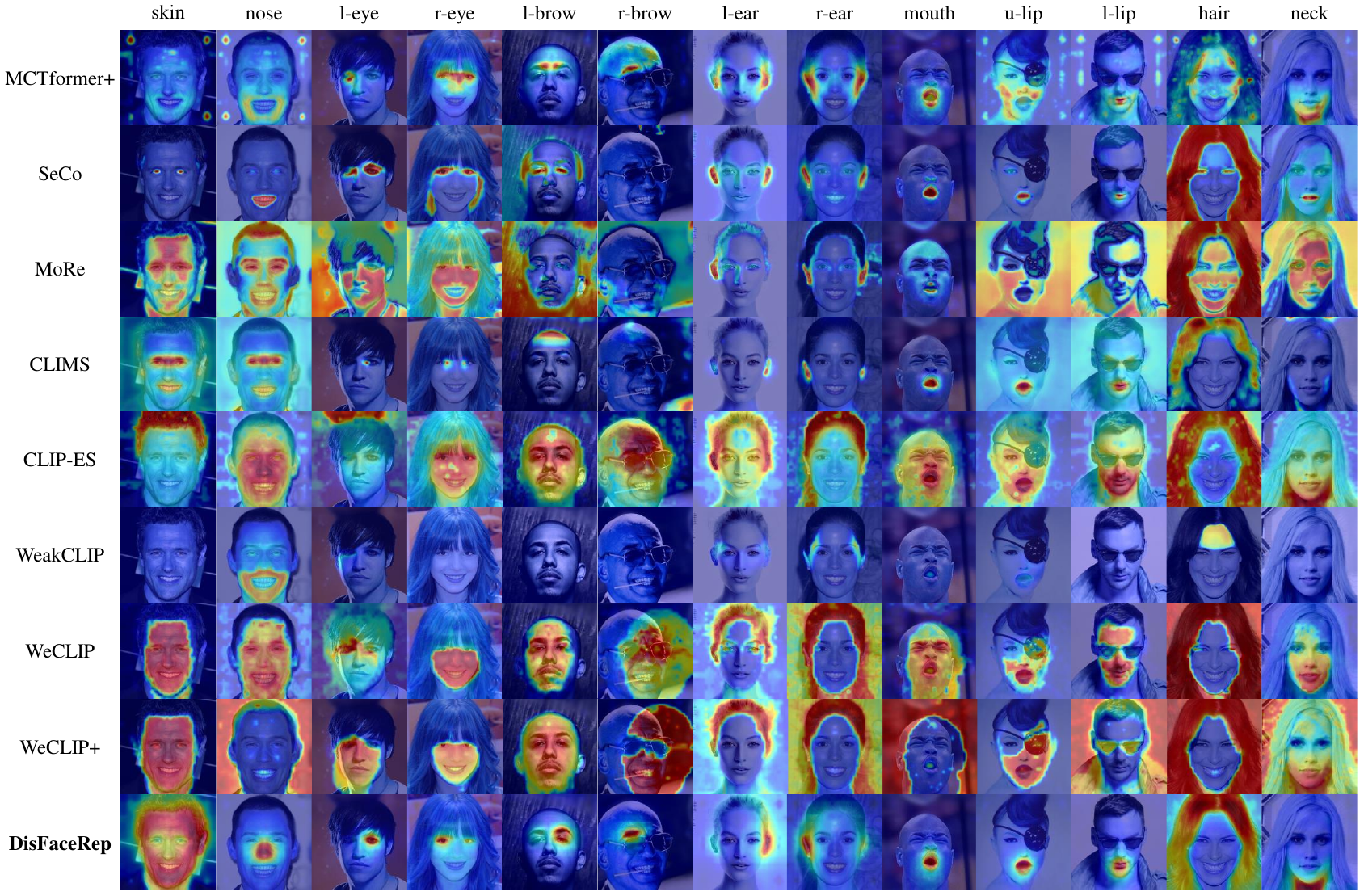}}
    \vspace{-5pt}
    \caption{Qualitative comparison of FCAMs for frequently co-occurring and visually similar facial components (left/right (l-/r-)eye/brow/ear, upper/lower (u-/l-)lip, skin, nose, mouth, hair, and neck) in the CelebAMask-HQ dataset.}
    \label{fig-celeba-cam}
    \vspace{-6pt}
\end{figure*}

\section{Experiments}
\subsection{Experimental Settings}
\subsubsection{Datasets and Evaluation Metrics}
We conduct experiments on three standard face parsing datasets: CelebAMask-HQ \cite{lee2020maskgan}, LaPa \cite{liu2020new}, and Helen \cite{smith2013exemplar}. CelebAMask-HQ contains 30,000 images, split into 24,183 for training, 2,993 for validation, and 2,824 for testing, with annotations for 19 classes, including background, skin, eyes, eyebrows, nose, mouth, lips, hair, ears, accessories, and clothing. To evaluate the performance of pseudo masks, we uniformly sampled 3,820 samples from the training set, forming a representative training subset. LaPa consists of 18,176 training images, 2,000 validation images, and 2,000 testing images, and is annotated for 11 classes, including background, skin, hair, nose, mouth, eyes, eyebrows, and lips. Helen contains 2,330 images with the same 11 labels as LaPa. It is split into 2,000, 230, and 100 images for training, validation, and testing. For evaluation, we report the mean F1 score across all facial categories, excluding the background.

\subsubsection{Implementation Details}
\label{Details}
Our proposed DisFaceRep adopts the Vision Transformer DeiT-S \cite{touvron2021training} with multi-class tokens as the backbone and initializes the model with ImageNet-pretrained weights \cite{deng2009imagenet}. Following the training setup in \cite{xu2024mctformer+}, we use the AdamW optimizer with an initial learning rate of $5 \times 10^{-6}$ and a batch size of 36. Input images are resized to $448 \times 448$. The loss weighting coefficients $[\alpha_0, \alpha_1, \beta_0, \beta_1, \beta_2, \lambda]$ are set to $[1.2, 0.2, 2.1, 0.1, 0.017, 0.05]$. During the pseudo-label generation phase, we remove Grounding DINO~\cite{liu2025grounding} and FLIP~\cite{li2024flip}, and follow the \cite{xu2024mctformer+} method to obtain FCAMs. Then, we set the threshold $\theta$ to 0.5 to generate pseudo-labels. For the face parsing model, we adopt DeepLab-V1 \cite{liang2015semantic} with ResNet38 \cite{wu2019wider} as the backbone. Notably, we omit CRF post-processing \cite{chen2014semantic} during inference. Additionally, CLIP~\cite{radford2021learning} is replaced with FLIP for all CLIP-based WSSS methods for fair comparisons.

\subsection{Comparison with WSSS Methods}
\subsubsection{Evaluation of Pseudo Labels and Face Parsing}
The first column of Tab.~\ref{tab-celeba} reports the mean F1 scores of pseudo masks generated by DisFaceRep and recent WSSS baselines on the CelebAMask-HQ training subset. Our method achieves the highest score of 44.81\%, significantly outperforming all competitors, including those using image-level supervision and those combining image-level and language supervision. Compared to SeCo \cite{yang2024separate}, the strongest image-level WSSS baseline, DisFaceRep achieves a gain of 9.78\%. Furthermore, it surpasses the best language-guided baseline, CLIP-ES \cite{lin2023clip}, by a large margin of 20.24\%, demonstrating the effectiveness of the proposed DisFaceRep in generating high-quality pseudo labels.

\begin{table}[!t]
    \caption{Mean F1 scores (\%) of pseudo masks and face parsing results generated by the proposed method and existing WSSS methods on the CelebAMask-HQ training subset, validation, and testing sets. Back. denotes the backbone network. Sup. denotes the weak supervision type.}
    \vspace{-5pt}
    \label{tab-celeba}
    \centering
    \setlength{\tabcolsep}{3.5pt}
    \begin{tabular}{ccccccc}
        \hline
        Methods & Back. & Sup. & Mask & val & test \\
        \hline
        MCTformer+ \cite{xu2024mctformer+} {\scriptsize \textcolor{gray}{[TPAMI24]}}                &ViT-B   &I    &29.89    &25.73  &25.88  \\
        SeCo \cite{yang2024separate} {\scriptsize \textcolor{gray}{[CVPR24]}}                       &ViT-B   &I    &35.03    &24.31  &24.64  \\
        MoRe \cite{yang2024more} {\scriptsize \textcolor{gray}{[AAAI25]}}                       &ViT-B   &I    &20.03    &7.92   &8.23   \\
        \hline
        CLIMS \cite{xie2022clims} {\scriptsize \textcolor{gray}{[CVPR22]}}                      &Res38  &I+L  &17.71    &15.06  &14.72  \\
        CLIP-ES \cite{lin2023clip} {\scriptsize \textcolor{gray}{[CVPR23]}}                    &Res101  &I+L  &24.57    &26.89  &27.48  \\ 
        WeakCLIP \cite{zhu2024weakclip} {\scriptsize \textcolor{gray}{[IJCV24]}}                   &ViT-B   &I+L  &21.78    &20.07  &19.74  \\
        WeCLIP \cite{zhang2024frozen} {\scriptsize \textcolor{gray}{[CVPR24]}}                     &ViT-B   &I+L  &22.67    &19.73  &20.37  \\
        WeCLIP+ \cite{zhang2025frozen} {\scriptsize \textcolor{gray}{[TPAMI25]}}                   &ViT-B   &I+L  &18.55    &18.18  &18.29  \\
        \rowcolor{gray!30}DisFaceRep {\scriptsize \textcolor{gray}{[Ours]}}      &ViT-B   &I+L  &\textbf{44.81} &\textbf{40.44} &\textbf{40.73}  \\
        \hline
    \end{tabular}
    \vspace{-10pt}
\end{table}

\begin{figure*}[t]
    \centering
    {\includegraphics[width=0.92\linewidth]{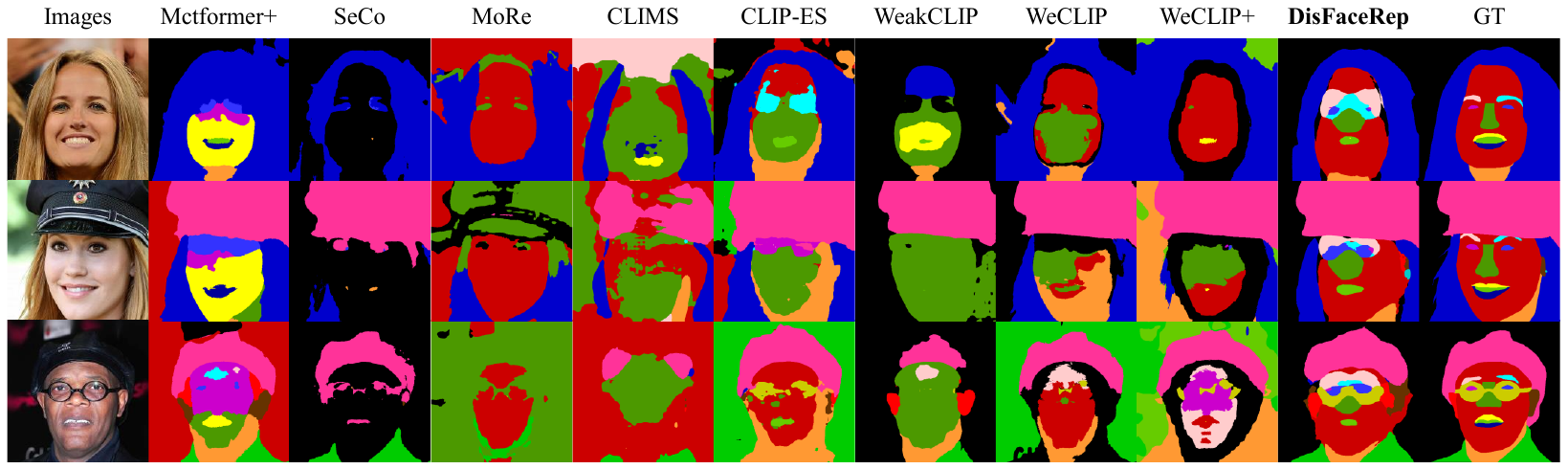}}
    \vspace{-4pt}
    \caption{Qualitative comparison of face parsing results on CelebAMask-HQ between WSSS methods and DisFaceRep.}
    \label{fig-celeba-seg}
    \vspace{-6pt}
\end{figure*}

\begin{figure}[t]
    \centering
    {\includegraphics[width=0.85\linewidth]{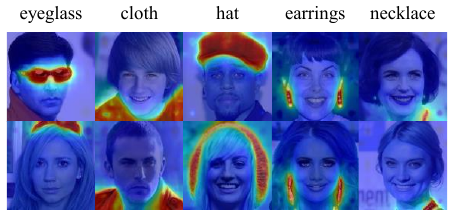}}
    \vspace{-4pt}
    \caption{Qualitative comparison of FCAMs for less frequently co-occurring and visually distinctive facial components in the CelebAMask-HQ dataset.}
    \label{fig-other-cam}
    \vspace{-12pt}
\end{figure}

Fig.~\ref{fig-celeba-cam} compares the FCAMs generated by DisFaceRep with those from conventional WSSS methods for facial components exhibiting strong co-occurrence and visual similarity. DisFaceRep consistently activates more accurate and complete regions for challenging components such as skin and nose, as shown in the first two columns of Fig.\ref{fig-celeba-cam}. In contrast, WSSS-based methods often miss these regions (e.g., SeCo \cite{yang2024separate}, WeakCLIP \cite{zhu2024weakclip}), underestimate them (e.g., MCTformer+ \cite{xu2024mctformer+}), or falsely activate unrelated areas (e.g., MoRe \cite{yang2024more}, CLIP-ES \cite{lin2023clip}). Moreover, WSSS methods also struggle to distinguish symmetrical components such as left/right eyes, eyebrows, and ears—frequently activating them uniformly or mislocalizing them. DisFaceRep, on the other hand, maintains strong component-level discrimination. Fig.~\ref{fig-other-cam} further demonstrates that DisFaceRep also produces accurate activation for less frequently co-occurring attributes, including eyeglass, cloth, hat, earrings, and necklace.

We further evaluate our method by training a segmentation network—DeepLab-V1 \cite{liang2015semantic} with ResNet38 \cite{wu2019wider}—on the CelebAMask-HQ training set. The face parsing results, shown in the second and third columns of Tab.~\ref{tab-celeba}, demonstrate that DisFaceRep achieves the best performance among all compared WSSS methods, with mean F1 scores of 40.44\% on the validation set and 40.73\% on the test set. Compared to MCTformer+, which uses only image-level supervision, DisFaceRep yields improvements of 14.71\% and 14.85\% on the validation and test sets, respectively. It also outperforms CLIMS, which incorporates both image-level and language supervision, by 25.38\% and 26.01\%. Qualitative results in Fig.~\ref{fig-celeba-seg} further confirm that DisFaceRep produces more accurate and compact segmentation maps than prior WSSS methods. While full activation of left and right attributes remains a challenge, our method significantly mitigates the co-occurrence issue.

\subsubsection{Cross-Dataset Face Parsing}
To assess cross-dataset generalization, we evaluate models trained on CelebAMask-HQ pseudo labels using the LaPa \cite{liu2020new} and Helen \cite{smith2013exemplar} datasets. It is noted that the 19-category CelebAMask-HQ covers all 11 categories in LaPa and Helen. Hence, except for 8 extra classes, we can map the same 11 categories from CelebAMask-HQ to LaPa and Helen by indices (e.g., nose label: 2 $\Rightarrow$ 6). Additionally, these datasets contain no overlapping images, preventing data leakage and ensuring valid cross-dataset evaluation. As shown in Tab.~\ref{tab_cross}, DisFaceRep consistently achieves the highest performance across both datasets. On LaPa, it obtains 28.90\% and 28.79\% mean F1 scores on the validation and test sets, respectively, which is more than 10\% higher than the next best method, CLIP-ES. On Helen, DisFaceRep also outperforms all baselines, reaching 26.83\% on the validation set and 27.66\% on the test set. These results demonstrate the strong cross-dataset generalization ability of DisFaceRep.

\begin{table}[!t]
    \caption{Cross-dataset face parsing results on the LaPa and Helen datasets. Mean F1 scores (\%) are reported on the validation and test sets.}
    \vspace{-3pt}
    \label{tab_cross}
    \centering
    \begin{tabular}{ccccc}
        \hline
        \multirow{2}{*}{Methods} & \multicolumn{2}{c}{LaPa \cite{liu2020new}} & \multicolumn{2}{c}{Helen \cite{smith2013exemplar}} \\
        \cline{2-3} \cline{4-5}
        & val & test & val & test \\
        \hline
        MCTformer+ \cite{xu2024mctformer+} {\scriptsize \textcolor{gray}{[TPAMI24]}} &11.30  &11.08  &12.14  &13.12  \\
        SeCo \cite{yang2024separate} {\scriptsize \textcolor{gray}{[CVPR24]}}        &10.32  &10.40  &10.67  &11.21  \\
        MoRe \cite{yang2024more} {\scriptsize \textcolor{gray}{[AAAI25]}}        &5.34   &5.39   &6.13   &7.06   \\
        \hline
        CLIMS \cite{xie2022clims}  {\scriptsize \textcolor{gray}{[CVPR22]}}       &6.60   &6.38   &6.35   &7.01   \\
        CLIP-ES \cite{lin2023clip} {\scriptsize \textcolor{gray}{[CVPR23]}}     &18.44  &18.50  &19.01  &19.60  \\
        WeakCLIP \cite{zhu2024weakclip} {\scriptsize \textcolor{gray}{[IJCV24]}}    &5.98   &5.70   &6.75   &7.69   \\
        WeCLIP \cite{zhang2024frozen} {\scriptsize \textcolor{gray}{[CVPR24]}}      &12.81  &12.85  &13.14  &13.29  \\
        WeCLIP+ \cite{zhang2025frozen} {\scriptsize \textcolor{gray}{[TPAMI25]}}    &14.09  &13.92  &12.15  &12.65  \\
        \rowcolor{gray!30}DisFaceRep {\scriptsize \textcolor{gray}{[Ours]}} &\textbf{28.90}  &\textbf{28.79}  &\textbf{26.83}  &\textbf{27.66}  \\
        \hline
    \end{tabular}
    \vspace{-6pt}
\end{table}

\begin{table}[!t]
    \caption{Ablation studies of component disentanglement modules of CCD and TCD on CelebAMask-HQ training subset. The results above are based on pseudo labels. Base. denotes the baseline model.}
    \vspace{-3pt}
    \label{tab_dis}
    \centering
    \setlength{\tabcolsep}{2.5pt}
    \begin{tabular}{ccccccc}
        \hline
        Methods        &skin &nose &r-eye &r-brow &l-ear &Mean F1 \\
        \hline
        Base.          &14.65 &10.68 &1.15 &11.42 &6.00 &29.67 \\
        Base.+CCD      &30.87 &47.41 &3.71 &10.20 &6.80 &33.86 \\
        Base.+CCD+TCD  &\textbf{72.68} &\textbf{73.14} &\textbf{22.55} &\textbf{19.31} &\textbf{26.65} &\textbf{44.81} \\
        \hline
    \end{tabular}
    \vspace{-6pt}
\end{table}

\subsection{Ablation Studies and Analysis}
\subsubsection{Effectiveness of CCD and TCD Modules}
We conduct ablation studies on the CelebAMask-HQ training subset to evaluate the effectiveness of the proposed CCD and TCD modules for pseudo label generation. As shown in Tab.~\ref{tab_dis}, the baseline model without CCD and TCD modules achieves a mean F1 score of 29.67\%. Incorporating CCD alone improves the score to 33.86\%,  demonstrating the effectiveness of suppressing frequently co-occurring facial components to reduce feature entanglement and improve activation quality. Adding TCD further boosts performance to 44.81\%, demonstrating the effectiveness of text-guided component disentanglement. We also report the results for frequently co-occurring facial attributes (e.g., skin, nose, right eye, right eyebrow, and left ear). The results show that combining CCD and TCD in DisFaceRep significantly improves prediction accuracy for these challenging components.

\begin{table}[!t]
    \caption{Comparison of the quality of pseudo labels using different combinations of loss functions. The mean F1 scores are reported on the CelebAMask-HQ training subset.}
    \vspace{-4pt}
    \label{tab_loss}
    \centering
    \begin{tabular}{cccccccc}
        \hline
        $L_{pos}^{cls}$ &$L_{pos}^{ffc}$ &$L_{neg}^{cls}$ &$L_{neg}^{ffc}$ &$L_{neg}^{bfc}$ &$L_{reg}$ & Mean F1 \\ 
        \hline
        \checkmark & & & & & &35.01 \\
        \checkmark & \checkmark & & & & &36.37 \\
        \checkmark & \checkmark & \checkmark & & & &37.49 \\
        \checkmark & \checkmark & \checkmark & \checkmark & & &41.17 \\
        \checkmark & \checkmark & \checkmark & \checkmark &\checkmark & &41.21 \\
        \checkmark & \checkmark & \checkmark & \checkmark &\checkmark &\checkmark &\textbf{44.81} \\
        \hline
    \end{tabular}
    \vspace{-6pt}
\end{table}

\begin{table}[!t]
    \caption{Comparison of model complexity and training efficiency with image size $448 \times 448$.}
    \vspace{-4pt}
    \label{tab_com}
    \centering
    \setlength{\tabcolsep}{2.5pt}
    \begin{tabular}{cccccc}
        \hline
        Methods &\makecell{Train\\Params}↓ &\makecell{Total\\Params}↓ &\makecell{Train\\Time}↓ &MACs↓ &FPS↑ \\
        \hline
        SeCo \cite{yang2024separate} &94.9 M &189.6 M &8.2 h &67.6 G &44.2 \\
        WeCLIP \cite{zhang2024frozen} &16.2 M &155.6 M &4.4 h &45.27 G &106.1 \\
        MCTformer+ \cite{xu2024mctformer+} &22.0 M &22.0 M &2.3 h &17.3 G &159.5 \\
        DisFaceRep &22.5 M &172.1 M &4.3 h &17.3 G &142.4 \\
        \hline
    \end{tabular}
    \vspace{-6pt}
\end{table}

\subsubsection{Impact of Loss Functions on TCD}
We conduct an ablation study on the CelebAMask-HQ training subset to evaluate the effectiveness of each loss function in the TCD module. As shown in Tab.~\ref{tab_loss}, using only the class-text positive loss $\mathcal{L}_{\text{pos}}^{\text{cls}}$ yields a baseline mean F1 score of 35.01\%. This already surpasses the baseline using only image-level supervision (29.67\% in Tab.\ref{tab_dis}), demonstrating the benefit of text-guided supervision in WSFP. Adding the foreground-level positive loss $\mathcal{L}_{\text{pos}}^{\text{ffc}}$ further improves the mean F1 score to 36.37\%, showing that both positive terms contribute to more accurate component activation. Introducing the class-text negative loss $\mathcal{L}_{\text{neg}}^{\text{cls}}$ raises performance to 37.49\%, while including all three negative losses ($\mathcal{L}_{\text{neg}}^{\text{cls}}$, $\mathcal{L}_{\text{neg}}^{\text{ffc}}$, and $\mathcal{L}_{\text{neg}}^{\text{bfc}}$) boosts the score to 41.21\%. Finally, adding the regularization term $\mathcal{L}_{\text{reg}}$ results in the highest performance of 44.81\%, indicating that enforcing spatial compactness in activation maps further enhances pseudo label quality. These results highlight the complementary roles of each loss and validate the effectiveness of the full TCD loss design in generating accurate and discriminative FCAMs.

\subsubsection{Sensitivity and Complexity Analysis}
\begin{figure}[t]
    \centering
    \includegraphics[width=0.92\linewidth]{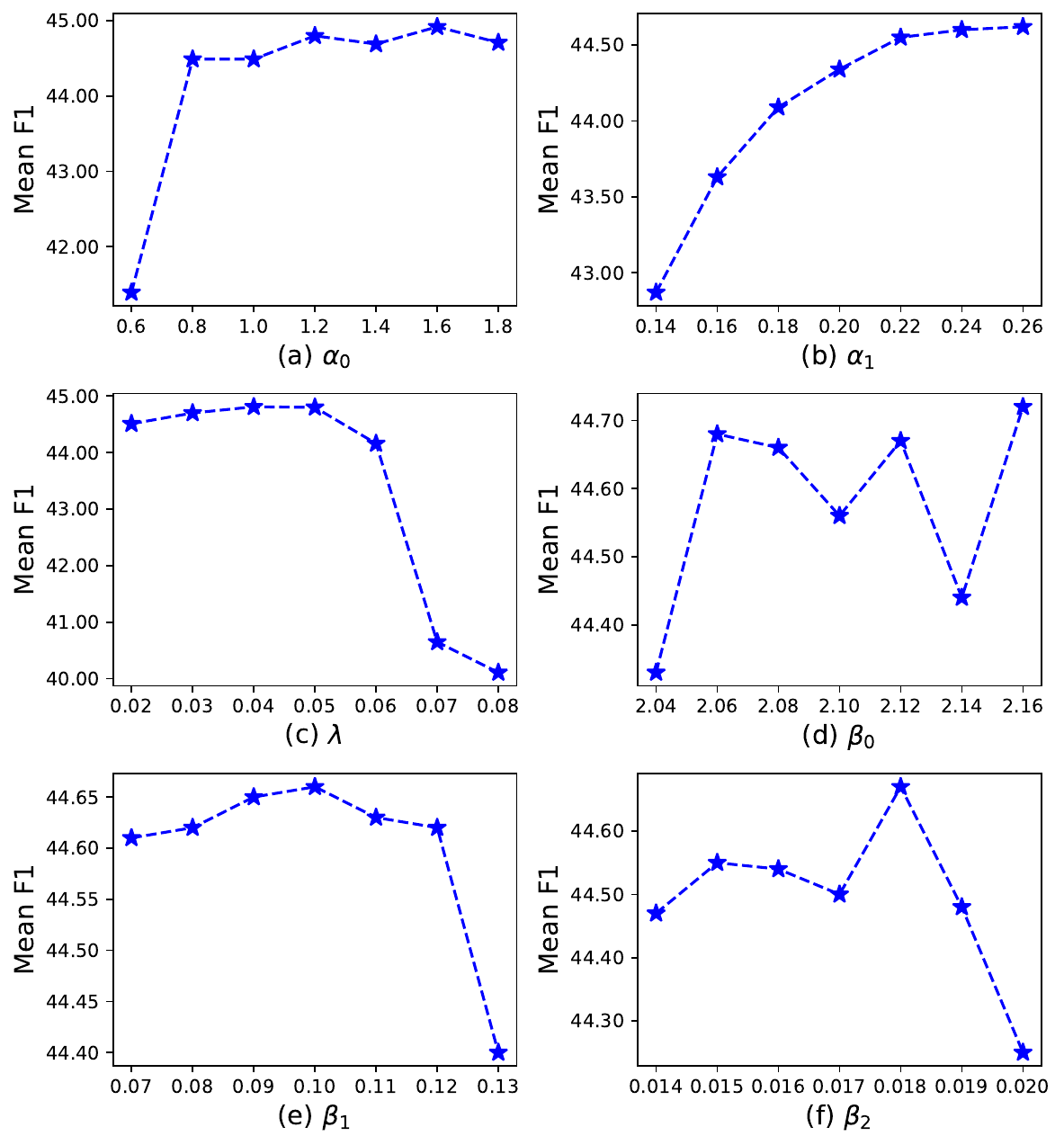}
    \vspace{-4pt}
    \caption{Sensitivity analysis on $\alpha_0$, $\alpha_1$, $\lambda$, $\beta_0$, $\beta_1$, and $\beta_2$. Mean F1 scores for pseudo-labels are reported on the CelebAMask-HQ training subset.}
    \label{fig_hyper}
    \vspace{-12pt}
\end{figure}

We conduct sensitivity analysis on hyperparameters in DisFaceRep: $\alpha_0$, $\alpha_1$, $\beta_0$, $\beta_1$, $\beta_2$, and $\lambda$, corresponding to Eq.\ref{eq: pos}, Eq.\ref{eq: neg}, and Eq.~\ref{eq: all}, using pseudo label generation on the CelebAMask-HQ training subset. As shown in Fig.\ref{fig_hyper}(a) and (b), the mean F1 score remains stable as $\alpha_0$ varies from 0.8 to 1.8 and $\alpha_1$ from 0.20 to 0.26, indicating robustness to the positive alignment weights. In Fig.\ref{fig_hyper}(c), varying $\lambda$ between 0.02 and 0.06 has minimal effect, confirming the stability of the regularization term. Fig.~\ref{fig_hyper}(d)–(f) show that the mean F1 scores remain consistently around 44\% as $\beta_0$, $\beta_1$, and $\beta_2$ vary within [2.06, 2.16], [0.07, 0.12], and [0.014, 0.019], respectively. These results show that DisFaceRep is robust to hyperparameter choices across a wide range of values.

Tab.~\ref{tab_com} shows the model complexity and training efficiency of our DisFaceRep, SeCo \cite{yang2024separate}, WeCLIP \cite{zhang2024frozen}, and MCTformer+ \cite{xu2024mctformer+}. It is worth noting that Grounding DINO \cite{liu2025grounding} is used before training, and FLIP \cite{li2024flip} is frozen in training and removed during inference in DisFaceRep. Therefore, our DisFaceRep has similar complexity and computational costs in terms of trainable and total parameters, training time, MACs, and FPS compared to the comparison methods.

\subsubsection{Impact of Language-Image Pre-training Models}
We evaluate the impact of different vision-language pretraining models, namely CLIP \cite{radford2021learning} and FLIP \cite{li2024flip}, on DisFaceRep. Using CLIP results in a mean F1 score of 37.20\% for pseudo-labeling on the CelebAMask-HQ training subset, while FLIP achieves a significantly higher score of 44.81\%. This 7.61\% improvement demonstrates the effectiveness of domain-specific pretraining. As FLIP is fine-tuned on facial image-text pairs, it provides stronger semantic alignment for facial components, leading to more accurate pseudo label generation. These results highlight the importance of choosing a vision-language model that is well-aligned with the task domain.

\section{Conclusion}
We introduced Weakly Supervised Face Parsing (WSFP), a new task that focuses on segmenting facial components using only weak supervision. WSFP presents unique challenges due to the frequent co-occurrence and high visual similarity of facial attributes. To address this, we proposed DisFaceRep, a representation disentanglement framework that explicitly suppresses dominant components via co-occurring component disentanglement and implicitly enhances component separation through text-guided alignment. Extensive experiments demonstrate that DisFaceRep generates more complete and compact activation maps compared to existing weakly supervised semantic segmentation methods. Cross-dataset evaluations further confirm its strong generalization ability, validating the effectiveness of disentangling co-occurring facial components for robust weakly supervised face parsing.

\begin{acks}
This work was supported by the National Natural Science Foundation of China under Grants 82261138629 and 62206180, Guangdong Basic and Applied Basic Research Foundation under Grant 2023A1515010688, Guangdong Provincial Key Laboratory under Grant 2023B1212060076, Shenzhen Municipal Science and Technology Innovation Council under Grant JCYJ20220531101412030, and XJTLU Research Development Funds under Grant RDF-23-01-053.
\end{acks}

\bibliographystyle{ACM-Reference-Format}
\balance
\bibliography{ref}

\end{document}